\documentclass{applemlr}

\usepackage{amsmath}
\usepackage{enumerate}
\usepackage{algorithm}
\usepackage{algorithmic}
\usepackage{amsfonts}
\usepackage{amsthm}
\usepackage{cleveref}
\usepackage{colortbl}
\usepackage{amssymb}
\usepackage{xspace}
\usepackage{adjustbox}
\usepackage{booktabs}
\usepackage{mathtools}
\usepackage{enumitem}
\usepackage{silence}
\usepackage{dsfont}
\usepackage[table,dvipsnames]{xcolor}
\usepackage{multirow}
\usepackage{makecell}
\usepackage{float}

\makeatletter
\patchcmd{\wrong@fontshape}{\@gobbletwo}{}{}{}
\makeatother
\numberwithin{equation}{section}

\makeatletter
\AtBeginDocument{
  \urlstyle{sf}
  
}
\makeatother

\crefname{assumption}{assumption}{assumption}
\Crefname{assumption}{Assumption}{Assumptions}

\usepackage[textsize=tiny]{todonotes}

\usepackage{tikz}
\usetikzlibrary{decorations.pathreplacing, arrows.meta}

\theoremstyle{plain}

\theoremstyle{definition}

\theoremstyle{remark}

\title{Stochastic KV Routing: Enabling Adaptive Depth-Wise Cache Sharing}

\author{Anastasiia Filippova}
\author{David Grangier}
\author{Marco Cuturi}
\author{João Monteiro}

\affiliation{Apple}

\abstract{Serving transformer language models with high throughput requires caching Key-Values (KVs) to avoid redundant computation during autoregressive generation. The memory footprint of KV caching is significant and heavily impacts serving costs. This work proposes to lessen these memory requirements. While recent work has largely addressed KV cache reduction via compression and eviction along the temporal axis, we argue that the \emph{depth} dimension offers an orthogonal and robust avenue for optimization.
Although prior research suggests that a full cache for every layer is redundant, implementing cross-layer cache sharing remains a practical challenge; existing methods typically suffer from reduced throughput or increased time-to-first-token.
In this paper, we demonstrate that dropping a layer's cache offers efficient optimization without information loss. We propose a simple training approach: random cross-layer attention. During training, layers randomly choose to attend either to their own KV states or those of a preceding layer. This stochastic process adapts the model to be robust to various depth-wise cache sharing strategies, ensuring flexibility for unknown hardware constraints at deployment time.
Our evaluations show that applying this scheme during pre-training or fine-tuning enables depth-wise cache sharing for various model families. Furthermore, for larger models in data-constrained settings, this approach is suggestive of a regularization-like effect, frequently preserving or improving performance while significantly reducing the cache's memory footprint.
}

\metadata[Correspondence]{\sffamily
Anastasiia Filippova: \url{a_filippova@apple.com};
David Grangier: \url{grangier@apple.com};
Marco Cuturi: \url{m_cuturi@apple.com};
João Monteiro: \url{jmonteiro2@apple.com}.
}

\begin{document}

\maketitle

\section{Introduction}\label{sec:intro}

Deploying large-scale language models requires careful inference optimizations to navigate the delicate trade-off between quality of user experience (latency or time to first token, throughput) and serving costs. KV caching avoids redundant computations in autoregressive generation at the expense of a large memory footprint. The KV cache scales linearly with batch size, sequence length, and model depth; its footprint outsizes the memory required for parameters in many practical settings with long inputs or outputs. The cache size drives inference cost and puts pressure on GPU memory, constraining the maximum batch size and context length. It also introduces significant latency due to the memory bandwidth overhead of loading these large tensors.

\begin{figure}
    \centering
    \includegraphics[width=0.6\linewidth]{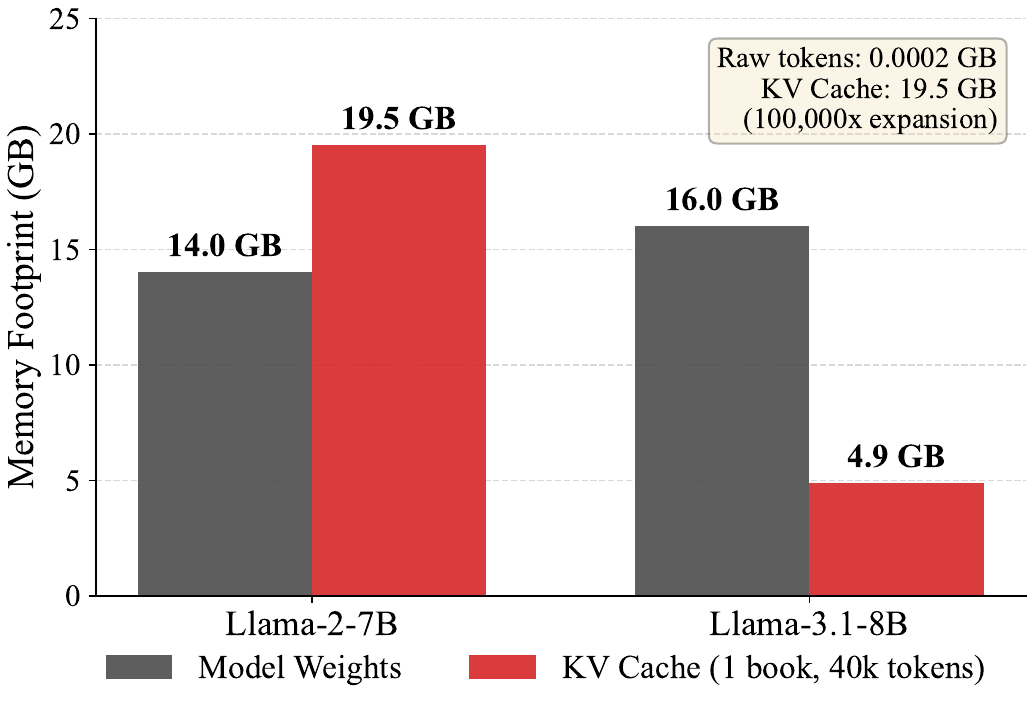}
    \caption{Language models are often described as extremely efficient databases, able to store training data in their parameters. The KV cache however doesn't reflect that, and a single book may require as much space as the model itself. Memory footprint per token can be obtained as follows for transformer decoders: \texttt{$2\times$\#layers$\times$\#kv\_heads$\times$\#head\_dim$\times2$}, where the leftmost factor of $2$ is due to storing both $K$ and $V$ representations, while the rightmost factor of $2$ is due to each floating point value requiring two bytes, assuming FP16 precision.}
    \label{fig:rel_cache_size}
\end{figure}

The inefficiency of the KV cache is particularly striking when one considers the fundamental nature of the data it represents. A language model takes as input a sequence of tokens; integers that typically require only 2 to 4 bytes each. Yet, to process these integer tokens, the model expands them into massive high-dimensional floating-point tensors at every single layer. For example, in \texttt{Llama-2-7B}~\citep{touvron2023llama}, a single token's integer representation is negligible, but its corresponding KV cache footprint is approximately \textbf{512 KB}\footnote{\texttt{$2\times$\#layers$\times$\#kv\_heads$\times$\#head\_dim$\times2$}, assuming FP16 precision.}. This represents a data expansion of over \textbf{100,000$\times$}. Similarly, as illustrated in Figure~\ref{fig:rel_cache_size}, later iterations of the model (\textit{i.e.}, \texttt{Llama-3-8B}~\citep{dubey2024llama}) require 128 KB per token. To put this into more tangible terms: caching the context of a book such as \emph{Alice's Adventures in Wonderland}~\cite{carroll1865alice} would consume roughly $1.4$ times the memory of the \texttt{Llama-2-7B} model weights themselves. It seems paradoxical that models, which are ostensibly compression machines that encode vast amounts of training data into their parameters, require such a large amount of memory to manage a single runtime context. Other more modern models such as \texttt{Qwen3-8B}~\citep{yang2025qwen3} improved relative to models released a few years ago, and require roughly 144 KB per token. This improvement is due to KV sharing within layers via approaches such as grouped query attention (GQA)~\citep{ainslie2023gqa}. Here, we claim more aggressive KV sharing can be carried out \emph{across layers} to induce even smaller caches.

The cache size grows since Key ($K$) and Value ($V$) states are stored for every token at every layer. To mitigate this, recent work has focused largely on \emph{temporal eviction} that identify and drop unimportant tokens along the time axis~\citep{zhang2023h2o, li2024snapkv, cai2024pyramidkv}. While effective, these strategies must contend with query-dependent token importance. A token deemed irrelevant for the current query might become crucial for the next, presenting a challenge where the model must either re-compute lost states or risk hallucinations~\citep{zhu2025oraclekv}. Determining which tokens to drop also induces additional computational overhead.

We argue that the \textbf{depth dimension} offers a robust, yet underutilized avenue for optimization, \emph{orthogonal} to temporal eviction and architectural variations. Recent work suggests that models exhibit high inter-layer redundancy, and a full per-layer cache is likely unnecessary~\citep{monteiro2024xc, wu2024layer}. However, using a single shared cache for all layers requires a separate encoder~\citep{monteiro2024xc}, which introduces cache update overhead as adding to the cache requires a full encoder forward pass, or estimating the top layers' output~\citep{wu2024layer}, which in turn significantly slows down pre-filling (\textit{i.e.}, building the cache), as multiple forward passes are required to reach reasonable approximation accuracy. \emph{Post-hoc} approaches were also proposed to enable cache sharing across layers~\citep{liu2024minicache}, but sharing is done only to a limited extent as models are trained with layer-specific Keys and Values.

To bridge this gap, we propose a training approach that breaks the rigid dependency between a layer and its own specific KV states during fine-tuning. We do so by means of performing cross-layer attention (CLA)~\citep{brandon2405reducing}, but randomly deciding on source previous layers that will provide $K$ and $V$ states. Namely, during training, for every forward pass of a layer $l$ and with probability $p$, the layer performs standard self-attention using its own $K$ and $V$ states (\textit{i.e.}, $l' = l$). With probability $1-p$, it is forced to attend to the $K$ and $V$ states of a randomly selected preceding layer $l'$ (where $l' < l$). This stochastic process simulates cache faults at a structural level. The layer learns to extract necessary information not just from its specific feature space, but from the general semantic representations available in earlier layers.

This approach, dubbed R-CLA, allows us to define flexible \emph{cache sharing strategies} at test time. For instance, we might choose to cache only every $4^{th}$ layer, forcing the intermediate 3 layers to reuse the nearest cached states. Because the model was trained against a randomized variety of such reuse patterns, it copes with different cache sharing strategies picked at testing time. Crucially, this flexibility allows a single model to be deployed across diverse hardware environments, from high-end clusters retaining 100\% of the cache to edge devices retaining only a fraction of it, without the need to train separate models for each observed hardware constraint.

In summary, our contributions are as follows:
\begin{itemize}
    \item We demonstrate that the disparity between token representation and cache size leads to a massive utilization gap, and that full layer-wise caching is redundant for maintaining prediction quality.
    \item Random Cross-Layer Attention (R-CLA): We introduce a simple, randomized training scheme that adapts models to be robust against depth-wise cache sharing. We show this method allows for memory reduction (\textit{e.g.}, dropping 50-75\% of layers) outperforming equivalent self-attention models.
    \item We observe that for larger models, the randomness introduced during training is suggestive of a regularization effect. Performance at full retention is frequently preserved or improved compared to standard full-cache baselines across multiple tasks.
\end{itemize}

\begin{figure}[t]
    \centering
    \includegraphics[width=0.7\linewidth]{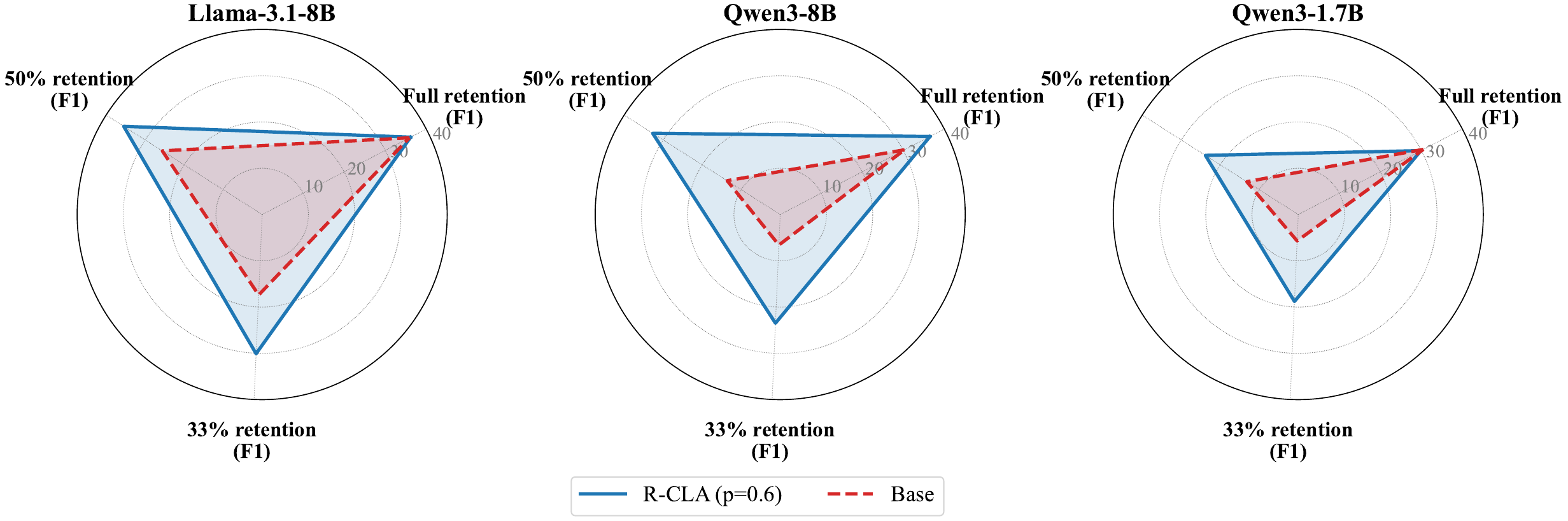}
    \caption{Performance comparison under varying cache retention rates for base models and their counterparts trained with random cross-layer attention. We report F1 scores for Question-Answer evaluations under different levels of cache retention for standard and R-CLA fine-tuned models. R-CLA avoids much of the degradation base models incur upon cache sharing, and preserves or improves performance at full cache.}
    \label{fig:radar_charts}
\end{figure}

\section{Related work}\label{sec:related_work}

The potential impact of the size of the KV cache in scaling inference has sparked a vast body of literature, primarily focused on the temporal dimension, architectural modifications, and, more recently, depth-wise redundancy.

\paragraph{Temporal Eviction and Compression.} 
The majority of research to date has focused on \emph{temporal eviction} or compression, trying to reduce the cache size by dropping or compressing tokens along the time axis. Early efforts, like StreamLM~\citep{xiao2024efficient}, identified a recency bias in attention patterns, proposing to keep only a small sliding window of recent tokens alongside crucial ``attention sinks'' (initial tokens). However, this often discards relevant semantic information stored in the middle of a sequence. To address this, methods like SnapKV~\citep{li2024snapkv} use attention scores to identify and preserve important chunks of the cache. This idea was extended by methods like PyramidKV~\citep{cai2024pyramidkv}, CAKE~\citep{qincake}, and LAVa~\citep{shen122025lava}, which vary the cache budget across different layers and attention heads. Despite their prevalence, these methods are inherently \emph{query-dependent}~\citep{zhu2025oraclekv}; because the importance of a token changes based on the query, eviction must be re-computed dynamically. While some recent attempts like OracleKV~\citep{zhu2025oraclekv} and KVZip~\citep{kim2025kvzip} try to mitigate this by predicting future token relevance using past queries, or by using streaming clustering for sub-linear query scoring~\citep{hooper2025multipole}, they either risk information loss or maintain a full index in memory that offsets the communication gains. Furthermore, methods like H2O~\citep{zhang2023h2o} and FastGen~\citep{ge2023model} focus on cache updates during generation, yet they do not reduce the peak memory footprint during the pre-filling phase for long contexts.

\paragraph{Architectural Improvements and Efficiency.} 
One could argue that the most successful reductions in cache footprint have come from architectural shifts rather than post-hoc eviction or compression approaches. By changing how the model handles attention, past work has meaningfully reduced the number of $K$ and $V$ vectors that need to be stored. Multi-Query Attention (MQA)~\citep{raffel2020exploring} and Grouped-Query Attention (GQA)~\citep{ainslie2023gqa} achieve this by sharing KV states across multiple query heads within a single layer. The impact of these architectural choices is significant: as noted previously, \texttt{Llama-2-7B} requires $512$KB/token, whereas the more modern \texttt{Qwen3-8B} architecture~\citep{yang2025qwen3} reduces the cache memory footprint to $144$KB/token primarily through the use of GQA. Similarly, cross-layer attention (CLA)~\citep{brandon2405reducing} makes it so two neighbor layers share a single set of KV states. Other structural innovations include State Space Models (SSMs)~\citep{dao2024transformers,yang2025gated} and hybrid architectures like \texttt{Kimi-k1.5}~\citep{team2025kimi} and \texttt{Nemotron-Nano-9B-v2}~\citep{basant2025nvidia}, which interleave attention with SSM layers and require no cache in non-attention blocks. Similarly, local/global self-attention architectures have been defined where local layers attend only to a small neighborhood close to the query, requiring a smaller cache~\citep{beltagy2020longformer,wang2025rattention}. Additionally, KV quantization~\citep{hooper2024kvquant} has proven effective at reducing the numerical precision of the cache. Notably, our proposed depth eviction is orthogonal to these architectural improvements and can be combined with GQA or quantization for compounded gains.

\paragraph{Depth-Wise Cache Sharing.}
While less explored than the temporal axis, one may exploit the redundancy of representations across the model's layers. XC-Cache~\citep{monteiro2024xc} demonstrated that a shared cache across layers is feasible but requires expensive updates due to calls to an external bi-directional encoder. Similarly, Layer-Condensed KV Cache~\citep{wu2024layer} proposes a shared single-layer cache but suffers from a significant increase in time-to-first-token due to the need for sequential pre-filling or multiple forward passes. MiniCache~\citep{liu2024minicache} takes a post-hoc approach by merging pairs of late layers, but its gains are modest since the model is not trained to handle this type of weight sharing. More recently, \citet{wu2025systematic} provide a systematic study of cross-layer sharing configurations, evaluating various patterns but requiring pre-training from scratch for each. CommonKV~\citep{wang2025commonkv} takes a training-free approach, using SVD to share KV projection weights across adjacent layers. KVSharer~\citep{yang2024kvsharer} identifies shareable layers via a dissimilarity metric without retraining, achieving moderate compression (~30\%). LISA~\citep{mu2024crosslayer} uses lightweight feed-forward networks and low-rank matrices to approximate cross-layer attention weight differences, requiring post-hoc training of alignment networks. We also note that our approach is mechanically distinct from structured dropout~\citep{fan2019reducing}, which skips entire layers during training. In R-CLA, every layer performs its full computation; only the source of the KV states is randomized.

These works provide strong evidence for the hypothesis that Transformer decoders can function with a shared set of KV states, much like the cross-attention mechanism in encoder-decoder models. Our work builds on this intuition by proposing a practical and efficient training scheme. Just as GQA shares KV states across \emph{heads}, we follow a similar approach to CLA and share them across \emph{layers}, though we use a randomized training scheme to ensure models remain robust to various cache sharing strategies. Unlike methods that require a specific fixed sharing pattern~\citep{wu2025systematic, mu2024crosslayer} or that operate post-hoc on unmodified models~\citep{yang2024kvsharer, wang2025commonkv}, R-CLA produces a single model adaptive to arbitrary sharing strategies at inference time.

\section{Enabling Cross-Layer Attention}\label{sec:x_layer_attention}

\subsection{Background}

Let a Transformer decoder consist of $L$ layers. For a given layer $l \in \{1, \dots, L\}$, the standard self-attention mechanism computes a query $Q_l$, a key $K_l$, and a value $V_l$ from the layer's input hidden states. The output is computed as $\text{Attn}(Q_l, K_l, V_l)$. In a standard inference setting, the pair $(K_l, V_l)$ constitutes the KV cache for layer $l$, which must be stored in memory to process future tokens.

\subsubsection{KV Cache Dynamics: Pre-fill and Sampling}
Inference in autoregressive language models proceeds in two distinct phases: \emph{pre-fill} where inputs are used to create a KV cache and \emph{sampling} where output tokens are generated and the cache updated. In more detail: 
\begin{itemize}
    \item \textbf{Pre-fill:} The model builds the KV cache by processing the initial prompt of $T$ tokens in parallel. For a model with $L$ layers and hidden dimension $H$, this operation generates $L \times T$ key-value pairs, which are written to memory. This initial cache establishes the context for generation, but as discussed previously, expands token representations quite significantly. 
    \item \textbf{Sampling:} The model generates tokens sequentially. For each new sampled token, the model computes its specific $K$ and $V$ representations at every layer and appends them to the existing cache.
\end{itemize}
Crucially, the cache size grows linearly with the sequence length. If $S$ is the size of the KV state for a single token at one layer, the total memory footprint $M$ after generating $t$ new tokens is given by $M = L \times (T + t) \times S$.

As the sequence length increases due to high $T$ or $t$, the memory bandwidth required to load this ever-growing cache becomes the primary bottleneck for inference speed. The bottleneck in standard inference is often not the computation itself, but the memory bandwidth required to load these large KV tensors from high bandwidth memory to the compute units (SRAM) at every layer, as illustrated in Algorithm~\ref{alg:standard_inf}.

\begin{algorithm}[H]
\caption{Standard Inference (Memory Bound)}\label{alg:standard_inf}
\begin{algorithmic}
\FOR{$l = 1$ \textbf{to} $L$}
    \STATE Compute Query $Q_l$
    \STATE \textbf{Load} $K_l, V_l$
    \STATE Compute $\text{Attn}(Q_l, K_l, V_l)$
    \STATE \textbf{Update} $K_l, V_l$
\ENDFOR
\end{algorithmic}
\end{algorithm}

\begin{figure}
    \centering
    \includegraphics[width=0.55\linewidth]{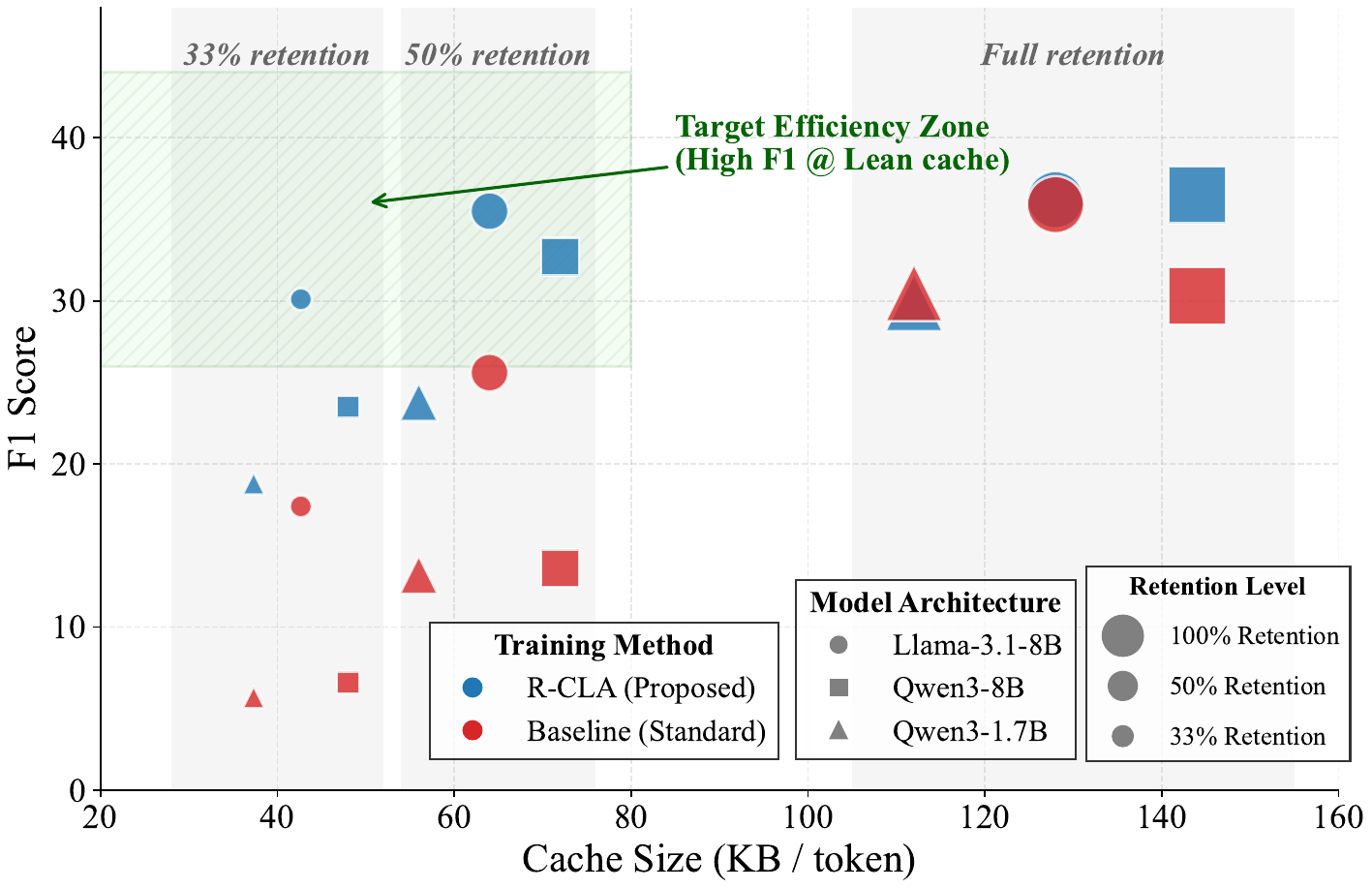}
    \caption{Cache size and prediction performance trade-offs for different models.}
    \label{fig:f1_cache_size_tradeoffs}
\end{figure}

\subsubsection{Cross-Layer Attention}
We define Cross-Layer Attention (CLA) as the operation where layer $l$ computes a \emph{fresh query} $Q_l$ based on its current input, but unlike the approach by~\citet{brandon2405reducing}, attends to any \emph{outdated keys and values} $(K_{l'}, V_{l'})$ generated by a preceding layer $l' < l$. The attention operation thus becomes:
\begin{equation}
    \text{Output}_l = \text{Attn}(Q_l, K_{l'}, V_{l'}) \quad \text{where } l' \leq l.
\end{equation}
When $l' = l$, this reduces to standard self-attention. When $l' < l$, the model reuses the cache of a previous layer, effectively bypassing the need to store or compute a unique cache for layer $l$.

\subsubsection{Cache Sharing Strategies}
We define a \emph{Cache Sharing Strategy} $\mathcal{S}$ as a subset of layer indices $\mathcal{S} \subseteq \{1, \dots, L\}$ that represent the layers authorized to maintain a KV cache in memory during inference. The goal is to minimize $|\mathcal{S}|$ while maintaining model performance.

Under a given strategy $\mathcal{S}$, for any layer $l$, the model uses the following mapping $\mu(l)$ to determine which cache to access:
\begin{equation}
    \mu(l) = 
    \begin{cases} 
    l & \text{if } l \in \mathcal{S} \\
    \max \{j \in \mathcal{S} \mid j < l\} & \text{if } l \notin \mathcal{S}.
    \end{cases}
\end{equation}
This mapping ensures that if a layer is not cached (\textit{i.e.}, $l \notin \mathcal{S}$), it defaults to re-using the cache of the nearest preceding layer that is still in memory. This enables highly efficient implementation patterns where the cache is loaded once and reused for multiple subsequent layer computations, as shown in Algorithm~\ref{alg:depth_eviction}, where \textbf{Load} and \textbf{Update} operations only happen for layers~$l$ such that $l \in \mathcal{S}$.

\begin{algorithm}[H]
\caption{Inference with Depth-Wise Cache Sharing}\label{alg:depth_eviction}
\begin{algorithmic}
\FOR{$l = 1$ \textbf{to} $L$}
    \STATE Compute Query $Q_l$
    \IF{$l \in \mathcal{S}$}
        \STATE \textbf{Load} $K_l, V_l$
        \STATE $\text{CurrentCache} \leftarrow K_l, V_l$
    \ENDIF
    \STATE $K_{l'}, V_{l'} \leftarrow \text{CurrentCache}$
    \STATE Compute $\text{Attn}(Q_l, K_{l'}, V_{l'})$
    \IF{$l \in \mathcal{S}$}
        \STATE \textbf{Update} $K_l, V_l$
    \ENDIF
\ENDFOR
\end{algorithmic}
\end{algorithm}

\subsection{Random Cross-Layer Attention Training}

A standard pre-trained model relies on specific feature alignments between $Q_l$ and $(K_l, V_l)$. Suddenly forcing $l \notin \mathcal{S}$ to attend to $\mu(l)$ at test time breaks these alignments, leading to degradation. To solve this, we must train the model to be invariant to the source of the keys and values.
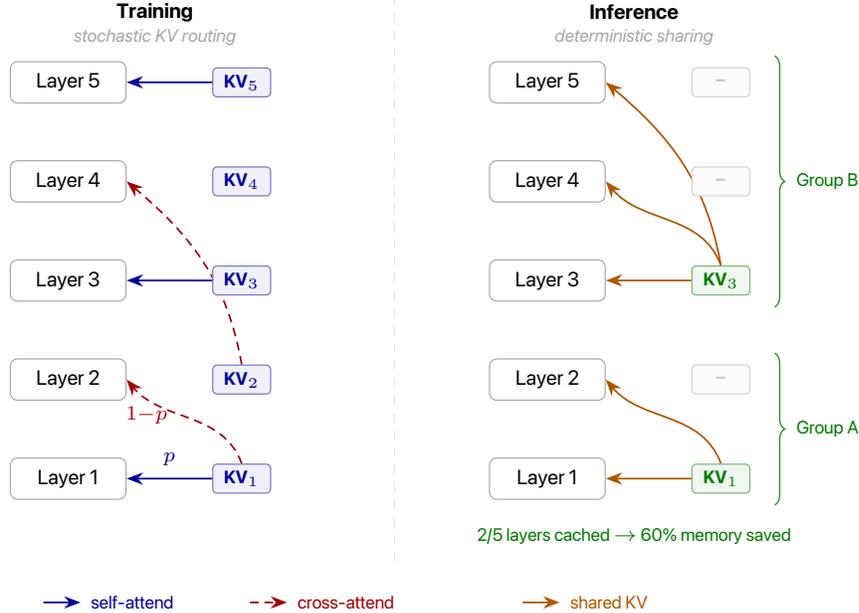
\begin{figure}[ht]
    \centering
    \resizebox{0.75\linewidth}{!}{\colorlet{rclablue}{blue!58!black}
\colorlet{rclared}{red!62!black}
\colorlet{rclagreen}{green!48!black}
\colorlet{rclaorange}{orange!68!black}
\colorlet{rclagray}{black!35}

\pgfdeclarelayer{bg}
\pgfsetlayers{bg,main}

\begin{tikzpicture}[
  >=Stealth,
  layer/.style={
    rectangle, rounded corners=2.5pt,
    minimum width=1.4cm, minimum height=0.5cm,
    draw=rclagray, fill=white, line width=0.4pt,
    font=\footnotesize\sffamily,
  },
  kvbox/.style={
    rectangle, rounded corners=1.5pt,
    minimum width=0.7cm, minimum height=0.35cm,
    line width=0.4pt,
    font=\scriptsize\sffamily\bfseries, inner sep=1pt,
  },
  kvself/.style  ={kvbox, draw=rclablue!55, fill=rclablue!6, text=rclablue},
  kvcached/.style={kvbox, draw=rclagreen!55, fill=rclagreen!6, text=rclagreen},
  kvempty/.style ={kvbox, draw=rclagray!50, fill=rclagray!3, text=rclagray!60},
  arrself/.style ={->, rclablue,   line width=0.6pt},
  arrcross/.style={->, rclared,    line width=0.6pt, densely dashed},
  arrshare/.style={->, rclaorange, line width=0.6pt},
  ptitle/.style={font=\footnotesize\sffamily\bfseries},
  psub/.style  ={font=\scriptsize\sffamily\itshape, text=rclagray},
  annot/.style ={font=\scriptsize\sffamily},
]

\pgfmathsetmacro{\vs}{1.2}    %
\pgfmathsetmacro{\kx}{2.1}    %
\pgfmathsetmacro{\px}{5.8}    %
\pgfmathsetmacro{\divx}{(\kx+\px)/2}  %

\node[ptitle] at (\kx/2, 4*\vs+0.85) {Training};
\node[psub]   at (\kx/2, 4*\vs+0.55) {stochastic KV routing};

\foreach \i [evaluate=\i as \y using {(\i-1)*\vs}] in {1,...,5} {
  \node[layer]  (TL\i) at (0,   \y) {Layer \i};
  \node[kvself] (TK\i) at (\kx, \y) {KV$_{\i}$};
}

\draw[arrself] (TK1.west) --
  node[midway, above=1pt, annot, text=rclablue] {$p$} (TL1.east);
\draw[arrself] (TK3.west) -- (TL3.east);
\draw[arrself] (TK5.west) -- (TL5.east);

\begin{pgfonlayer}{bg}
  \draw[arrcross] (TK1.north) to[out=110, in=-50]
    node[pos=0.55, left=2pt, annot, text=rclared] {$1{-}p$} (TL2.east);
  \draw[arrcross] (TK2.north) to[out=100, in=-40] (TL4.east);
\end{pgfonlayer}

\node[ptitle] at (\px+\kx/2, 4*\vs+0.85) {Inference};
\node[psub]   at (\px+\kx/2, 4*\vs+0.55) {deterministic sharing};

\foreach \i [evaluate=\i as \y using {(\i-1)*\vs}] in {1,...,5} {
  \node[layer] (IL\i) at (\px, \y) {Layer \i};
}

\node[kvcached] (IK1) at (\px+\kx, 0)       {KV$_1$};
\node[kvempty]  (IK2) at (\px+\kx, \vs)      {\textendash};
\node[kvcached] (IK3) at (\px+\kx, 2*\vs)    {KV$_3$};
\node[kvempty]  (IK4) at (\px+\kx, 3*\vs)    {\textendash};
\node[kvempty]  (IK5) at (\px+\kx, 4*\vs)    {\textendash};

\draw[arrshare] (IK1.west) -- (IL1.east);
\draw[arrshare] (IK3.west) -- (IL3.east);

\begin{pgfonlayer}{bg}
  \draw[arrshare] (IK1.north) to[out=110, in=-50] (IL2.east);
  \draw[arrshare] (IK3.north) to[out=110, in=-50] (IL4.east);
  \draw[arrshare] (IK3.north) to[out=100, in=-40] (IL5.east);
\end{pgfonlayer}

\draw[decorate, decoration={brace, amplitude=3.5pt, mirror},
      rclagreen!80, line width=0.5pt]
  (\px+\kx+0.65, -0.32) -- (\px+\kx+0.65, \vs+0.32)
  node[midway, right=4pt, annot, text=rclagreen] {Group A};
\draw[decorate, decoration={brace, amplitude=3.5pt, mirror},
      rclagreen!80, line width=0.5pt]
  (\px+\kx+0.65, 2*\vs-0.32) -- (\px+\kx+0.65, 4*\vs+0.32)
  node[midway, right=4pt, annot, text=rclagreen] {Group B};

\node[annot, rclagreen, align=center] at (\px+\kx/2, -0.7)
  {2/5 layers cached $\rightarrow$ 60\% memory saved};

\draw[rclagray!30, dashed, line width=0.3pt]
  (\divx, -1.0) -- (\divx, 4*\vs+1.1);

\draw[arrself]  (-0.3, -1.5) -- ++(0.45,0)
  node[right, annot, text=rclablue] {self-attend};
\draw[arrcross] (2.2, -1.5) -- ++(0.45,0)
  node[right, annot, text=rclared] {cross-attend};
\draw[arrshare] (\px-0.3, -1.5) -- ++(0.45,0)
  node[right, annot, text=rclaorange] {shared KV};

\end{tikzpicture}}
    \caption{Behavior of a model implementing R-CLA. During \emph{training}, each layer decides whether to attend to KV states obtained from the past layer's hidden states, or from a randomly chosen layer in the past, so each batch observes a different cache sharing strategy. During \emph{testing}, a fixed deterministic cache sharing scheme is employed, and the same group of layers always share the same KV cache.}
    \label{fig:rcla_training_testing}
\end{figure}
We propose \textbf{Random Cross-Layer Attention (R-CLA)}. Instead of training for a specific strategy $\mathcal{S}$, which would overfit the model to a fixed pattern, we introduce stochasticity during the forward pass as a means to make models robust against a large set of cache sharing strategies. Note that the current usage pattern of language models involves a fixed general-purpose model deployed to perform various tasks in many diverse serving settings, facing different hardware constraints. Ideally, a model would be able to cope with different levels of cache retention so as to enable serving under different conditions and hardware constraints. Not only that, as will be discussed with the evaluations as reported in Section~\ref{sec:eval}, randomness due to stochasticity in cross-layer attention is suggestive of a regularization-like effect, beneficial in the data-constrained fine-tuning setting.

During training, for every layer $l$ and every forward pass, we sample a decision variable $d \sim \text{Bernoulli}(p)$:
\begin{itemize}
    \item If $d=1$: The layer performs standard self-attention using $(K_l, V_l)$.
    \item If $d=0$: The layer performs cross-layer attention using $(K_{l'}, V_{l'})$, where $l'$ is sampled uniformly from $\{1, \dots, l-1\}$.
\end{itemize}

This scheme forces the query projection $Q_l$ to learn to interact with a wide variety of key/value distributions from previous layers. By explicitly breaking the dependency on the layer's own cache, we build robustness into the model itself. Consequently, at test time, we gain the flexibility to select different cache sharing strategies $\mathcal{S}$, as dictated by the specific constraints of the deployment environment, without needing to retrain the model.

We further illustrate the variations in behavior across training and testing in Figure~\ref{fig:rcla_training_testing}. During training, each layer decides whether to self-attend, \textit{i.e.}, ingest KV states obtained from the that past layer's output hidden states, or cross-layer attend to a previous layer picked at random. At testing time on the other hand, cache sharing schema are set deterministically. In the example depicted in Figure~\ref{fig:rcla_training_testing} for instance, the sharing scheme used at testing time is so that groups of two and then three layers share the cache, which reduces the overall cache memory footprint as well as the amount of data communication as a consequence.

The effect of R-CLA can be observed in Figure~\ref{fig:radar_charts} for different fine-tuned models on a Question-Answer setting that will be further discussed in Section~\ref{sec:eval}. Upon different levels of cache retention, obtained through different test caching strategies that keep only every $k$-th layer, models trained under R-CLA do withstand cache sharing and preserve base full cache performance to a greater extent than base models. Perhaps also relevant, performance at full cache is preserved or improved, which is suggestive of a regularization-like effect from R-CLA training. We note that this approach is mechanically distinct from structured dropout~\citep{fan2019reducing}, which skips entire layers during training to tolerate reduced depth at inference. In R-CLA, every layer performs its full computation; only the source of the KV states is randomized.

Similarly, in Figure~\ref{fig:f1_cache_size_tradeoffs}, we observe that effect even more clearly by looking at the tradeoffs between task performance in terms of F1 scores and the cache size, varying due to changing cache sharing strategies. R-CLA dominates standard self-attention in the Pareto sense, and preserves or improves base performance at a full cache.

\begin{figure}
    \centering
    \includegraphics[trim={0 0 0 1.1cm},clip,width=0.55\linewidth]{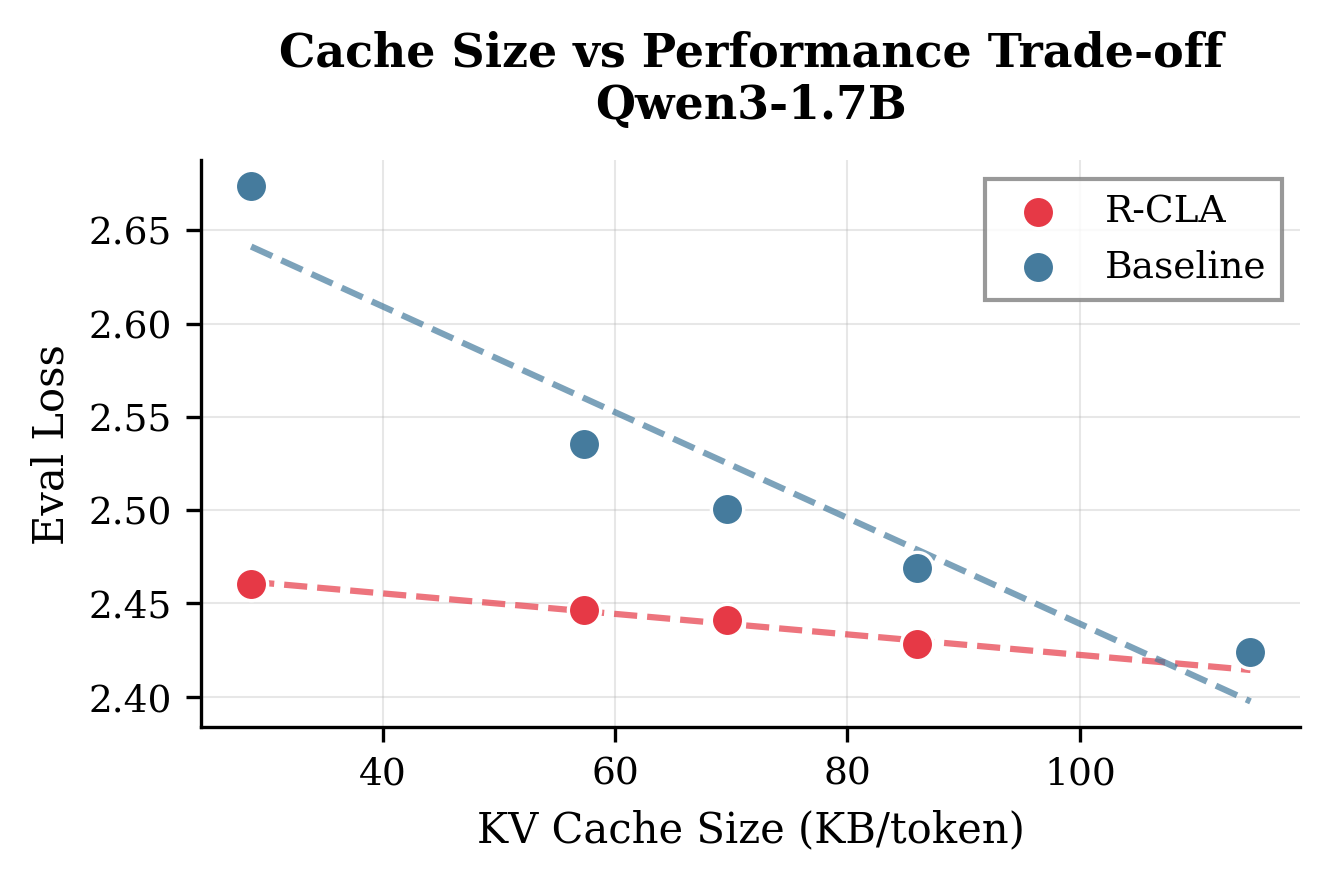}
    \caption{Cache size vs. evaluation loss for Qwen3-1.7B. R-CLA models (red) use the full $28$-layer architecture with depth-wise cache sharing, while baselines (blue) are shallower Transformers with the same cache footprint. Deeper models with shared cache consistently outperform shallower models, showing that cache sharing preserves the benefit of depth.}
    \label{fig:f5_pre_training}
\end{figure}

\section{Evaluation}\label{sec:eval}

We evaluate the impact of R-CLA in inference under cache sharing, and do so in two separate settings: compute-constrained pre-training in Section~\ref{sec:pretraining}, and data-constrained, task-specific supervised fine-tuning, in which case we focused on Question-Answer (QA) tasks, as discussed in Section~\ref{sec:finetuning}. While pre-training experiments assess whether R-CLA can be introduced early on and whether it impacts pre-training dynamics or induces instabilities, QA tasks are meaningful in our setting since performance critically depends on parsing and retaining information provided in the input prompt. As inference in our case targets sharing prompt data representations across model layers, possibly overriding relevant information, QA serves as the perfect proxy for assessing whether R-CLA renders shared cache models able to perform at parity with models that have access to the full context representation.

\subsection{Pre-training}
\label{sec:pretraining}

To verify that training under our cache sharing scheme does not destabilize pre-training, we train decoder-only Transformers from scratch both with and without R-CLA. 
For baseline comparison, we train shallower Transformers where the number of layers corresponds to the effective cache size under R-CLA. For instance, an R-CLA model with $p = 0.5$ (retaining $x$ layers' worth of cache at inference) is compared against a $x$-layer baseline Transformer. All models are trained using an equal number of tokens.
For R-CLA models, we use the full $2\times x$-layer architecture and vary the probability of attending to preceding layers.

\subsubsection{Experiment setup}

We use a \texttt{Qwen-1.7B}--style decoder-only Transformer \citep{yang2025qwen3} and pre-train it from scratch on a subset of the OpenWeb corpus~\citep{Gokaslan2019OpenWeb}. All runs use a fixed context length of 2{,}048 tokens. We optimize with AdamW ($\beta_1 = 0.9$, $\beta_2 = 0.99$)~\citep{loshchilov2017decoupled} with weight decay $0.1$, and apply gradient clipping with a maximum norm of $0.1$. The learning rate follows a linear warm-up to $1\times 10^{-4}$ over the first $5\%$ of training steps, followed by cosine decay for the remainder. All experiments are conducted on NVIDIA H100 GPUs. For a controlled comparison, all models are trained for an identical token budget of $34$B tokens, which is Chinchilla-optimal~\citep{hoffmann2022training} for a $1.7$B-parameter model.
For R-CLA we  vary $p \in \{0.25, 0.5, 0.6, 0.75\}$ and for the baseline number of hidden layers $\in \{7, 14, 17, 21, 28\}$.
  \begin{table}[ht]
  \centering
  \caption{Pre-training evaluation loss for Qwen3-1.7B trained with R-CLA at varying cross-layer attention probability $p$. Training remains stable across all values, with loss increasing by less than 2\% even at $p=0.75$.}
  \label{tab:pretraining_loss}
  \begin{tabular}{lccccc}
  \toprule
  $p$ & 0.00 & 0.25 & 0.50 & 0.60 & 0.75 \\
  \midrule
  \textbf{Eval Loss} & 2.424 & 2.428 & 2.441 & 2.446 & 2.461 \\
  \bottomrule
  \end{tabular}
  \end{table}

\subsubsection{Results}
Table \ref{tab:pretraining_loss} reports the evaluation loss for Qwen3-1.7B models trained with varying R-CLA probability $p$. Training remains stable across all configurations: the evaluation loss increases from $2.42$ at $p = 0$ (standard training) to only $2.46$ at $p = 0.75$, representing less than a $2\%$ degradation despite three-quarters of attention operations being redirected to preceding layers during training. This confirms that R-CLA does not destabilize pre-training dynamics. In Appendix~\ref{sec:training_curves}, we show that training is stable under R-CLA. 
\begin{table}[t]
\centering
\caption{Effect of R-CLA on F1 Score for models fine-tuned for Question-Answer. We compare the baseline standard self-attention with R-CLA trained under $p=0.6$. Evaluation is performed across three depth-wise cache retention levels.}
\label{tab:full_results}
\resizebox{\linewidth}{!}{%
\begin{tabular}{cl ccc ccc ccc}
\toprule
 & & \multicolumn{3}{c}{\textbf{Llama-3.1-8B}} & \multicolumn{3}{c}{\textbf{Mistral-7B}} & \multicolumn{3}{c}{\textbf{Qwen3-8B}} \\
\cmidrule(lr){3-5} \cmidrule(lr){6-8} \cmidrule(lr){9-11}
\textbf{Dataset} & \textbf{Retention} & \textbf{Base} & \textbf{R-CLA} & \textbf{$\Delta\%$} & \textbf{Base} & \textbf{R-CLA} & \textbf{$\Delta\%$} & \textbf{Base} & \textbf{R-CLA} & \textbf{$\Delta\%$} \\
\midrule
\multirow{3}{*}{HotpotQA}
 & 100\% & 0.203 & \textbf{0.306} & +51.1 & 0.215 & \textbf{0.242} & +12.4 & 0.233 & \textbf{0.357} & +53.1 \\
 & 50\%  & 0.171 & \textbf{0.305} & +78.7 & 0.162 & \textbf{0.211} & +30.0 & 0.085 & \textbf{0.314} & +267.8 \\
 & 25\%  & 0.080 & \textbf{0.237} & +196.2 & 0.031 & \textbf{0.162} & +421.3 & 0.011 & \textbf{0.098} & +826.6 \\
\midrule
\multirow{3}{*}{MSMarco}
 & 100\% & 0.301 & \textbf{0.318} & +5.6 & 0.310 & \textbf{0.316} & +1.7 & 0.291 & \textbf{0.329} & +13.0 \\
 & 50\%  & 0.226 & \textbf{0.324} & +43.2 & 0.236 & \textbf{0.300} & +26.9 & 0.112 & \textbf{0.227} & +102.4 \\
 & 25\%  & 0.047 & \textbf{0.217} & +358.2 & 0.069 & \textbf{0.180} & +161.8 & 0.027 & \textbf{0.058} & +110.6 \\
\midrule
\multirow{3}{*}{RepLiQA}
 & 100\% & \textbf{0.655} & 0.649 & -0.9 & \textbf{0.649} & 0.648 & -0.2 & 0.441 & \textbf{0.478} & +8.4 \\
 & 50\%  & 0.314 & \textbf{0.597} & +90.0 & 0.291 & \textbf{0.547} & +88.2 & 0.154 & \textbf{0.349} & +126.8 \\
 & 25\%  & 0.073 & \textbf{0.307} & +322.3 & 0.084 & \textbf{0.214} & +154.7 & 0.029 & \textbf{0.065} & +126.4 \\
\midrule
\multirow{3}{*}{SQuAD v2}
 & 100\% & 0.583 & \textbf{0.758} & +30.0 & 0.573 & \textbf{0.750} & +30.8 & 0.705 & \textbf{0.728} & +3.3 \\
 & 50\%  & 0.427 & \textbf{0.740} & +73.2 & 0.388 & \textbf{0.654} & +68.4 & 0.269 & \textbf{0.627} & +132.9 \\
 & 25\%  & 0.257 & \textbf{0.628} & +144.6 & 0.078 & \textbf{0.500} & +541.7 & 0.042 & \textbf{0.284} & +575.0 \\
\midrule
\multirow{3}{*}{TriviaQA}
 & 100\% & 0.328 & \textbf{0.360} & +9.8 & 0.205 & \textbf{0.216} & +5.6 & 0.146 & \textbf{0.295} & +101.7 \\
 & 50\%  & 0.290 & \textbf{0.351} & +20.8 & 0.138 & \textbf{0.158} & +14.6 & 0.032 & \textbf{0.248} & +671.9 \\
 & 25\%  & 0.137 & \textbf{0.291} & +112.2 & 0.032 & \textbf{0.121} & +273.9 & 0.005 & \textbf{0.131} & +2287.3 \\
\bottomrule
\end{tabular}%
}
\end{table}

Figure \ref{fig:f5_pre_training} presents the cache size versus evaluation loss trade-off, comparing two approaches under equivalent cache budgets: (1) R-CLA models using the full $28$-layer architecture with depth-wise cache sharing, and (2) baseline models with fewer layers but full per-layer caching. Across all cache sizes tested, R-CLA models consistently achieve lower evaluation loss than their shallower baseline counterparts. This demonstrates that depth with cache sharing is preferable to reduced depth under identical inference memory constraints.
\begin{table}[t]
\centering
\caption{Ablations on top of Llama-3.1-8B. We compare R-CLA ($p=0.6$) against deterministic CLA fixed at every $2^{nd}$ or $4^{th}$ layer, and their randomly applied counterparts (RD-CLA, $p=0.6$).}
\label{tab:ablation}
\begin{tabular}{cl ccccc}
\toprule
\multirow{2}{*}{\textbf{Dataset}} & \multicolumn{1}{c}{\multirow{2}{*}{\textbf{Retention}}} & \multicolumn{5}{c}{\textbf{Method}} \\
\cmidrule(lr){3-7}
 & & \textbf{R-CLA} & \textbf{CLA@2} & \textbf{CLA@4} & \textbf{RD-CLA@2} & \textbf{RD-CLA@4} \\
\midrule
\multirow{3}{*}{HotpotQA}
 & 100\% & 0.306 & 0.192 & \textbf{0.331} & 0.223 & 0.167 \\
 & 50\%  & \textbf{0.305} & 0.149 & 0.284 & 0.179 & 0.134 \\
 & 25\%  & \textbf{0.237} & 0.089 & 0.172 & 0.100 & 0.071 \\
\midrule
\multirow{3}{*}{MSMarco}
 & 100\% & 0.318 & 0.282 & 0.303 & \textbf{0.323} & 0.300 \\
 & 50\%  & \textbf{0.324} & 0.184 & 0.159 & 0.247 & 0.211 \\
 & 25\%  & \textbf{0.217} & 0.065 & 0.085 & 0.062 & 0.076 \\
\midrule
\multirow{3}{*}{RepLiQA}
 & 100\% & 0.649 & 0.520 & 0.476 & \textbf{0.652} & 0.621 \\
 & 50\%  & \textbf{0.597} & 0.237 & 0.162 & 0.298 & 0.270 \\
 & 25\%  & \textbf{0.307} & 0.087 & 0.085 & 0.075 & 0.076 \\
\midrule
\multirow{3}{*}{SQuAD v2}
 & 100\% & 0.758 & 0.633 & \textbf{0.768} & 0.660 & 0.630 \\
 & 50\%  & \textbf{0.740} & 0.387 & 0.456 & 0.521 & 0.521 \\
 & 25\%  & \textbf{0.628} & 0.290 & 0.397 & 0.314 & 0.313 \\
\midrule
\multirow{3}{*}{TriviaQA}
 & 100\% & 0.360 & 0.332 & \textbf{0.376} & 0.357 & 0.357 \\
 & 50\%  & \textbf{0.351} & 0.280 & 0.321 & 0.326 & 0.315 \\
 & 25\%  & \textbf{0.291} & 0.145 & 0.165 & 0.170 & 0.158 \\
\bottomrule
\end{tabular}
\end{table}

\subsection{Fine-tuning}
\label{sec:finetuning}
We performed task-specific fine-tuning to evaluate whether (i) cache sharing can be induced efficiently on pre-trained models, (ii) R-CLA enables depth-wise cache sharing, and (iii) any regularization effect is induced by R-CLA. We target tasks that require models to retrieve knowledge from the input context since that would test whether cache sharing preserves information. We thus fine-tune a variety of models of different sizes both with and without R-CLA. All models train on the same data for 50{,}000 steps under a batch size of 128 and a maximum input length of 8{,}192 tokens. Like pre-training, we train with AdamW but use $\beta_1 = 0.9$, and $\beta_2 = 0.95$. Weight decay is set to $0.1$. The learning rate is warmed-up linearly to $5\times 10^{-6}$ over the first $1.5\%$ of training steps, and decays linearly to $0$.

\subsubsection{Data Preparation}

We curated a dataset from five sources to cover a diverse range of reasoning requirements. This includes HotpotQA~\citep{yang2018hotpotqa}, which requires multi-hop reasoning to synthesize answers from different parts of the context; SQuAD v2~\citep{rajpurkar2018know}; MSMarco~\citep{bajaj2016ms}; and TriviaQA~\citep{joshi2017triviaqa}. Additionally, we included RepLiQA~\citep{monteiro2024repliqa}, a dataset consisting of fictional content. This ensures that the evaluation strictly measures the model's ability to find answers within the context, rather than relying on parametric knowledge acquired during pre-training. The combined dataset consists of \texttt{(context, question, answer)} triples.%

To improve robustness, we applied specific data augmentation strategies. For all datasets, we randomized the order of the input components during training. With a probability of $0.5$, the format follows a \texttt{question -> context -> answer} structure; otherwise, it follows the standard \texttt{context -> question -> answer} format. While placing the question first theoretically benefits the model by allowing the query to condition the context encoding, this causal advantage is rarely available at test time. We therefore employ this randomization primarily as a means to diversify the training distribution.

Specific augmentation was applied to HotpotQA, where multiple context passages are provided, some of which are irrelevant confounders. For each training example, we generated three variations with different permutations of the concatenated context passages. This prevents the model from relying on position bias and encourages robust information retrieval across the input sequence.

\subsubsection{Results}

We evaluated our approach across three distinct model architectures: \texttt{Qwen3-8B}~\citep{yang2025qwen3}, \texttt{Mistral\-7B}~\citep{jiang2023mistral7b}, and \texttt{Llama\-3.1\-8B}~\citep{dubey2024llama}. To ensure a fair comparison, all models were trained with identical compute budgets and data mixtures, differing strictly in the application of the R-CLA mechanism versus standard self-attention. Results in Table~\ref{tab:full_results} demonstrate the impact of R-CLA on cache sharing tolerance. Across all cases, cross-layer attention enables cache sharing, and retaining less of the cache preserves performance to a much higher extent than in standard self-attention models.

In low-retention regimes (25\% retention), base models suffer catastrophic collapse, whereas R-CLA models retain substantial capabilities in some cases. Furthermore, R-CLA incurs no penalty in the standard full-retention setting, and frequently boosts performance (\textit{e.g.}, +50.7\% on HotpotQA for Llama-3.1). This suggests a regularization effect from stochastic training, beneficial in data-constrained settings.

\subsubsection{Ablation}

We conducted an ablation study using \texttt{Llama-3.1-8B} to assess the specific contribution of randomness in our approach versus the general benefits of sharing KV states. We compare our fully random method (R-CLA) against two baseline categories. First, CLA@k represents a deterministic fixed sharing scheme where groups of $k$ layers share KV states (\textit{e.g.}, layers $0$ to $k-1$ share one KV). Second, to disentangle the effect of stochastic training, we evaluate RD-CLA@k (Random-Deterministic CLA). This method uses the same fixed sharing scheme as CLA@k, but applies cross-layer attention stochastically: at each forward step at each layer, the model flips a coin to decide whether to attend to the fixed shared layer or its local KV.

The results in Table~\ref{tab:ablation} highlight distinct roles for sharing and stochasticity. In the no-sharing (100\% retention) regime, fixed CLA@k schemes frequently match or slightly exceed the performance of R-CLA. This suggests that the regularization benefits observed in previous sections stem primarily from the information bottleneck created by KV sharing, rather than randomness itself. However, under missing KV states (50\% and 25\% retention), the role of randomness becomes apparent, and adding stochasticity to fixed schemes (RD-CLA) provides improvements over deterministic baselines. Ultimately, the unstructured randomness of R-CLA consistently outperforms all other variations under missing states, indicating that exposing the model to a diverse set of sharing patterns during training helps learn representations that are resilient to missing cache.

\subsection{Inference Efficiency}
\label{sec:inference_efficiency}

To quantify the practical benefits of depth-wise cache sharing, we benchmark decode throughput, time-to-first-token (TTFT), KV cache memory, and peak GPU memory. We use a \texttt{Qwen3-8B}-scale architecture (36 layers, 4096 hidden, 32 attention heads, 8 KV heads, head\_dim=128) with bfloat16 precision on a single 80GB GPU. We sweep across input lengths from 512 to 32K tokens, sharing group sizes ($g \in \{1, 2, 4, 8\}$, where $g=1$ is the standard baseline), and batch sizes from 1 to 16. Under cache sharing with group size $g$, every $g$ consecutive layers share a single KV cache: the first layer in each group (the ``leader'') computes and stores $K, V$, while the remaining layers skip their $K, V$ projections and reuse the leader's cached states.

\begin{table}[t]
\centering
\caption{Inference efficiency across input lengths. Architecture matches \texttt{Qwen3-8B} (36 layers, 8 KV heads, bfloat16) on a single 80GB GPU. Group size $g{=}1$ is the standard baseline; $g{=}4$ shares every 4 consecutive layers. Time to first token (TTFT) measures prefill latency.}
\label{tab:benchmark_input_lengths}
\begin{tabular}{cc cccc}
\toprule
\textbf{Input} & \textbf{Group} & \textbf{Peak Mem.} & \textbf{KV Cache} & \textbf{TTFT} & \textbf{Throughput} \\
\textbf{Length} & \textbf{Size} & \textbf{(MB)} & \textbf{(MB)} & \textbf{(ms)} & \textbf{(tok/s)} \\
\midrule
\multirow{2}{*}{2{,}048}
 & 1 & 16{,}573 & 306   & 66   & 32.9 \\
 & 4 & 16{,}357 & 77    & 63   & 43.0 \\
\cmidrule{1-6}
\multirow{2}{*}{8{,}192}
 & 1 & 19{,}319 & 1{,}170 & 297  & 34.0 \\
 & 4 & 18{,}455 & 293   & 286  & 41.6 \\
\cmidrule{1-6}
\multirow{2}{*}{16{,}384}
 & 1 & 22{,}981 & 2{,}322 & 713  & 34.2 \\
 & 4 & 21{,}253 & 581   & 692  & 41.1 \\
\cmidrule{1-6}
\multirow{2}{*}{32{,}768}
 & 1 & 30{,}305 & 4{,}626 & 1{,}903 & 22.8 \\
 & 4 & 26{,}849 & 1{,}157 & 1{,}868 & 26.1 \\
\bottomrule
\end{tabular}
\end{table}

Table~\ref{tab:benchmark_input_lengths} reports results across input lengths at batch size 1. KV cache memory scales as $1/g$: at 8K tokens, it drops from 1170 MB (baseline) to 293 MB ($g{=}4$), a 4$\times$ reduction. Decode throughput improves consistently, from 34.0 tok/s (baseline) to 41.6 tok/s ($g{=}4$, +22\%) at 8K context, due to skipping $K$/$V$ projections on non-leader layers. Peak memory savings grow with context length: at 32K tokens, $g{=}4$ saves 3.5 GB versus baseline.

\begin{table}[t]
\centering
\caption{Batch size scaling at 8{,}192-token context. Same architecture as Table~\ref{tab:benchmark_input_lengths}. At batch size 16, the baseline ($g{=}1$) exceeds GPU memory, while cache sharing ($g{=}4$) completes successfully.}
\label{tab:benchmark_batch_scaling}
\begin{tabular}{cc cccc}
\toprule
\textbf{Batch} & \textbf{Group} & \textbf{Peak Mem.} & \textbf{KV Cache} & \textbf{TTFT} & \textbf{Throughput} \\
\textbf{Size} & \textbf{Size} & \textbf{(MB)} & \textbf{(MB)} & \textbf{(ms)} & \textbf{(tok/s)} \\
\midrule
\multirow{2}{*}{2}
 & 1 & 22{,}973 & 2{,}322 & 588   & 33.3 \\
 & 4 & 21{,}245 & 580    & 573   & 40.3 \\
\cmidrule{1-6}
\multirow{2}{*}{4}
 & 1 & 30{,}281 & 4{,}643 & 1{,}201 & 22.7 \\
 & 4 & 26{,}825 & 1{,}161 & 1{,}159 & 25.9 \\
\cmidrule{1-6}
\multirow{2}{*}{8}
 & 1 & 44{,}897 & 9{,}287 & 2{,}395 & 12.8 \\
 & 4 & 37{,}985 & 2{,}322 & 2{,}324 & 14.7 \\
\cmidrule{1-6}
\multirow{2}{*}{16}
 & 1 & \multicolumn{4}{c}{\textit{Out of Memory}} \\
 & 4 & 60{,}306 & 4{,}643 & 4{,}696 & 8.0 \\
\bottomrule
\end{tabular}
\end{table}

Table~\ref{tab:benchmark_batch_scaling} shows the effect of batch size scaling at 8K context. The benefits of cache sharing compound at larger batch sizes. Notably, at batch size 16, the baseline configuration runs out of memory, while and $g{=}4$ complete successfully, directly demonstrating that cache sharing enables higher serving capacity on the same hardware.

We note that these gains represent a conservative lower bound. Our implementation skips $K$/$V$ projections for non-leader layers and avoids allocating their caches, but the attention computation itself (loading $K$,$V$ from HBM) remains the same whether a layer reads from its own cache or a leader's. Backend-level optimizations could go further: for instance, keeping the shared $K$,$V$ in SRAM across consecutive layers, or fusing their attention computations into a single memory load.

\section{Conclusion}\label{sec:conclusion}

We addressed the KV cache memory bottleneck in language model inference. In line with previous work, we showed that the standard practice of maintaining a full cache for every layer is redundant and that models can effectively operate with significantly leaner caches by exploiting depth-wise correlations. By introducing random cross-layer attention (R-CLA), a simple yet effective randomized training strategy, we showed that pre-trained models can be adapted to be robust against various depth-wise cache sharing strategies, and using R-CLA early, during pre-training, can be done effectively. Our evaluations confirm that this approach reduces memory consumption, enabling longer contexts and larger batch sizes. We also observe that R-CLA is suggestive of a regularization-like effect for larger models under data-constrained settings: by breaking rigid layer-wise dependencies, R-CLA frequently preserves or improves performance at full retention. Looking forward, this work opens several avenues for optimizing inference \emph{orthogonally} to quantization or temporal pruning. While our method enables flexible eviction strategies, future work could explore adaptive strategies that dynamically adjust the cache depth based on the complexity of the input query or the current generation step. Additionally, investigating the interplay between depth-wise eviction and architectural innovations like Grouped-Query Attention (GQA) or State Space Models (SSMs) could yield further compounded efficiency gains.

\paragraph{Limitations.} Enabling R-CLA requires access to training resources; future research might investigate post-hoc methods or lightweight adapters that can induce similar cross-layer robustness without full parameter updates. Our experiments focus on compute-optimal training regimes and do not explore overtraining. We have not evaluated R-CLA on Mixture-of-Experts (MoE) architectures, though the method should be directly applicable since it only modifies the attention KV source. Our fine-tuning evaluation is focused on QA tasks; while these are a strong proxy for context retention under cache sharing, broader task evaluation would strengthen the conclusions. The composition of depth-wise cache sharing with temporal eviction methods and KV quantization is a natural extension that we leave to future work.

\bibliography{biblio}
\bibliographystyle{unsrtnat}

\clearpage
\appendix
\section{Training curves}
\label{sec:training_curves}

\begin{figure}[h]
    \centering
    \includegraphics[trim={0 0 0 0cm},clip,width=0.55\linewidth]{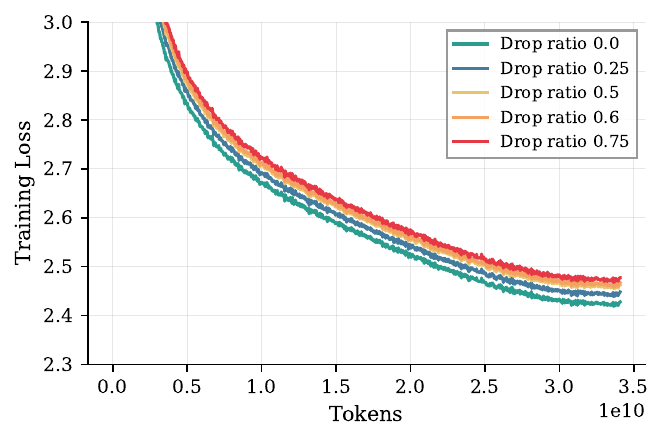}
    \caption{Training loss as a function of tokens processed for Qwen3-1.7B with varying KV cache sharing ratios.}
    \label{fig:pretraining_loss}
\end{figure}

In Figure~\ref{fig:pretraining_loss}, we show training curves for experiments reported in Section~\ref{sec:pretraining}. We pre-train a \texttt{Qwen-1.7B}--style decoder from scratch on a subset of the OpenWeb corpus~\citep{Gokaslan2019OpenWeb}. We train different models using different levels of cache sharing or dropping ratio. That is, we use R-CLA with $p \in \{0.25, 0.5, 0.6, 0.75\}$. All models are trained for an identical token budget of $34$B tokens. Training curves show that R-CLA incurs no training instability even under aggressive $p$, and hyperparameters transfer reasonably well from standard self-attention models to R-CLA ones. R-CLA can be introduced at pre-training time if cache sharing is desirable downstream.

\newpage
\section{Fine-tuning Training Dynamics}
\label{sec:finetuning_dynamics}

\begin{figure}[h]
    \centering
    \includegraphics[width=0.7\linewidth]{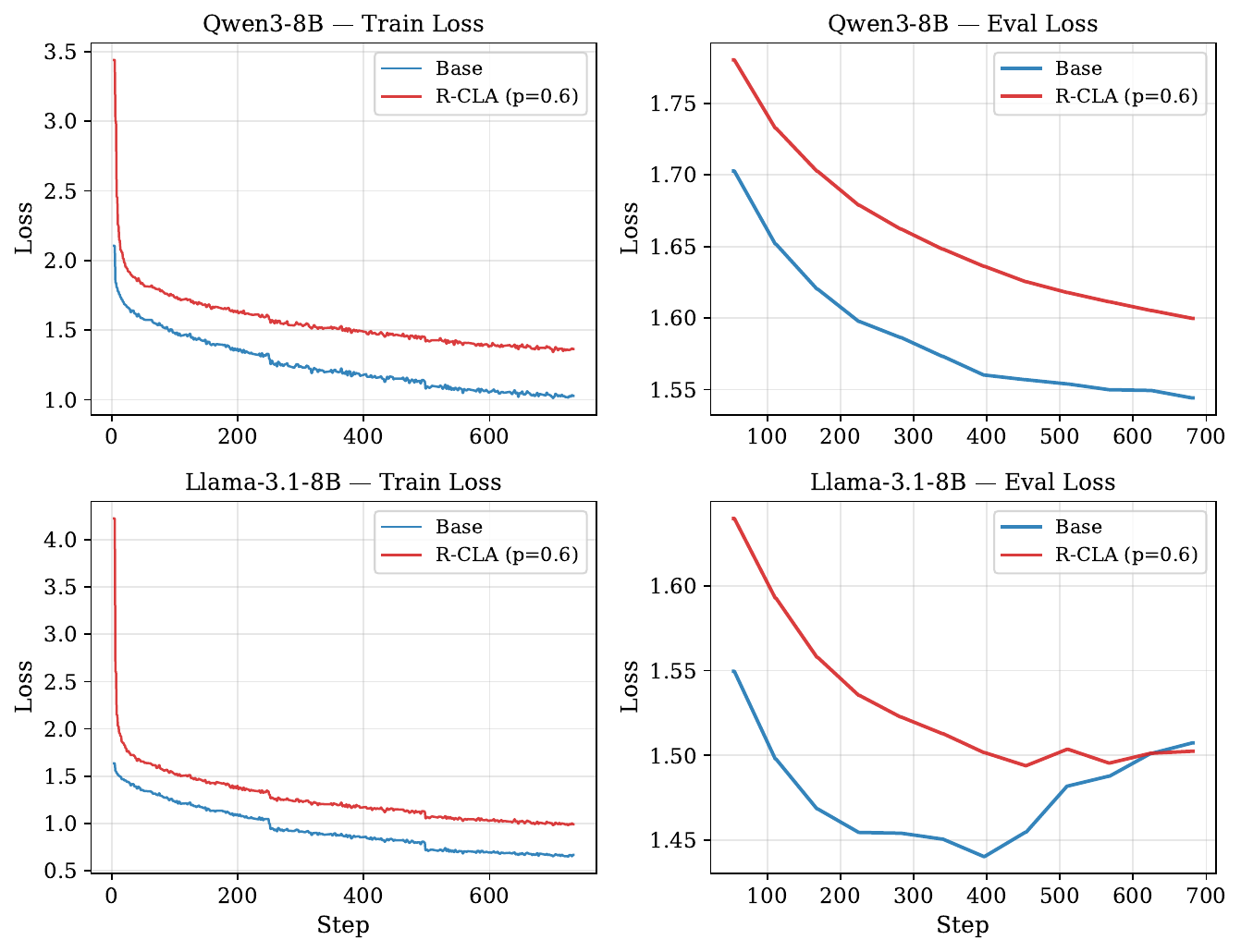}
    \caption{Train loss (left) and eval loss (right) over fine-tuning steps for Base (blue) vs.\ R-CLA (red) on \texttt{Qwen3-8B} (top) and \texttt{Llama-3.1-8B} (bottom). R-CLA shows slower learning throughout, consistent with a regularization-like effect from stochastic KV routing.}
    \label{fig:finetuning_dynamics}
\end{figure}

Figure~\ref{fig:finetuning_dynamics} plots train loss and eval loss over fine-tuning steps for Base vs.\ R-CLA on \texttt{Qwen3-8B} and \texttt{Llama-3.1-8B}. In both cases, stochastic KV routing slows training (higher train loss). For \texttt{Llama-3.1-8B}, the base model overfits while R-CLA delays the onset of overfitting. For \texttt{Qwen3-8B}, R-CLA learns consistently more slowly, in line with the pre-training results reported in Table~\ref{tab:pretraining_loss}. In a data-constrained fine-tuning setting, this means R-CLA models can train for more epochs before entering the overfitting regime.

\newpage
\section{R-CLA vs.\ CLA@k}
\label{sec:pareto_comparison}

\begin{figure}[h]
    \centering
    \includegraphics[width=0.9\linewidth]{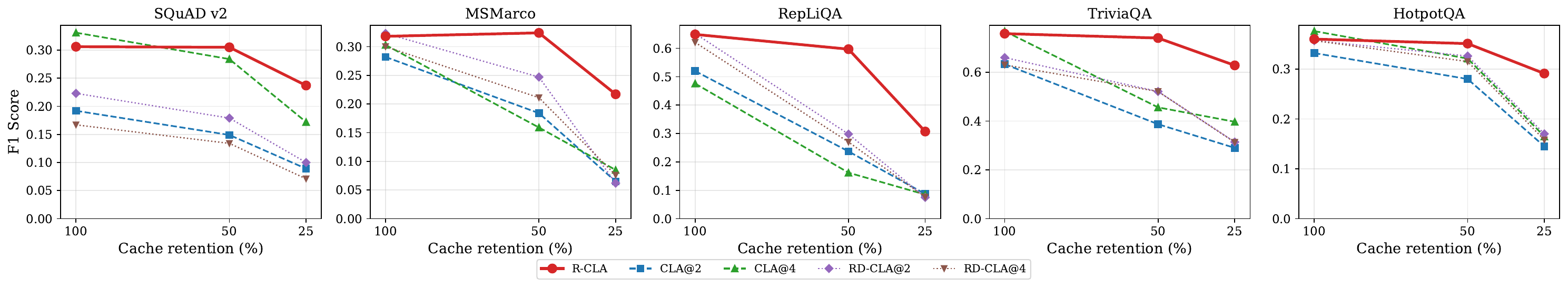}
    \caption{F1 score vs.\ cache retention (\%) for Llama-3.1-8B across five QA tasks. R-CLA (red, solid) maintains performance across retention levels, while CLA@$k$ variants (dashed) degrade outside their trained retention level. Data from Table~\ref{tab:ablation}.}
    \label{fig:pareto_overlay}
\end{figure}

Figure~\ref{fig:pareto_overlay} plots F1 score vs.\ cache retention for R-CLA, CLA@2, CLA@4, RD-CLA@2, and RD-CLA@4 across all five QA tasks using \texttt{Llama-3.1-8B}. R-CLA maintains competitive performance from 100\% to 25\% retention, while deterministic CLA@$k$ variants degrade sharply as retention moves away from their trained level. CLA@$k$ can match or exceed R-CLA at its specific trained retention level on some tasks (\textit{e.g.}, CLA@4 on TriviaQA at 100\%), but R-CLA is the only method that maintains competitive performance across all retention levels from a single training run.

\newpage
\section{Additional fine-tuning results}
\label{sec:fine_tuning_results_appendix}

\begin{table}[H]
\centering
\caption{We report Base, R-CLA ($p=0.6$), and relative improvement ($\Delta\%$) for F1, Exact Match (EM), and ROUGE-L across three cache retention levels.}
\label{tab:appendix_full_metrics}
\resizebox{\linewidth}{!}{%
\begin{tabular}{cccccccccccc}
\toprule
\multirow{2}{*}{\textbf{Dataset}} & \multirow{2}{*}{\textbf{Model}} & \multirow{2}{*}{\textbf{Retention}} & \multicolumn{3}{c}{\textbf{F1}} & \multicolumn{3}{c}{\textbf{Exact Match (EM)}} & \multicolumn{3}{c}{\textbf{ROUGE-L}} \\
\cmidrule(lr){4-6} \cmidrule(lr){7-9} \cmidrule(lr){10-12}
 & & & \textbf{Base} & \textbf{R-CLA} & \textbf{$\Delta\%$} & \textbf{Base} & \textbf{R-CLA} & \textbf{$\Delta\%$} & \textbf{Base} & \textbf{R-CLA} & \textbf{$\Delta\%$} \\
\midrule
\multirow{9}{*}{\rotatebox[origin=c]{90}{\textbf{HotpotQA}}} 
 & \multirow{3}{*}{Llama-3.1-8B} & 100\% & 0.203 & \textbf{0.306} & +51.1\% & 0.128 & \textbf{0.221} & +72.3\% & 0.206 & \textbf{0.307} & +49.0\% \\
 &  & 50\% & 0.171 & \textbf{0.305} & +78.7\% & 0.102 & \textbf{0.216} & +112.6\% & 0.173 & \textbf{0.305} & +75.8\% \\
 &  & 25\% & 0.080 & \textbf{0.237} & +196.2\% & 0.035 & \textbf{0.153} & +339.6\% & 0.088 & \textbf{0.240} & +171.6\% \\
 \cmidrule{2-12}
 & \multirow{3}{*}{Mistral-7B} & 100\% & 0.215 & \textbf{0.242} & +12.4\% & 0.122 & \textbf{0.141} & +15.8\% & 0.216 & \textbf{0.243} & +12.6\% \\
 &  & 50\% & 0.162 & \textbf{0.211} & +30.0\% & 0.081 & \textbf{0.106} & +30.5\% & 0.165 & \textbf{0.212} & +28.6\% \\
 &  & 25\% & 0.031 & \textbf{0.162} & +421.3\% & 0.014 & \textbf{0.078} & +467.3\% & 0.037 & \textbf{0.168} & +352.3\% \\
 \cmidrule{2-12}
 & \multirow{3}{*}{Qwen3-8B} & 100\% & 0.233 & \textbf{0.357} & +53.1\% & 0.157 & \textbf{0.273} & +73.3\% & 0.235 & \textbf{0.359} & +52.8\% \\
 &  & 50\% & 0.085 & \textbf{0.314} & +267.8\% & 0.024 & \textbf{0.227} & +863.8\% & 0.085 & \textbf{0.316} & +270.1\% \\
 &  & 25\% & 0.011 & \textbf{0.098} & +826.6\% & 0.000 & \textbf{0.041} & - & 0.014 & \textbf{0.108} & +692.5\% \\
\midrule
\multirow{9}{*}{\rotatebox[origin=c]{90}{\textbf{MSMarco}}} 
 & \multirow{3}{*}{Llama-3.1-8B} & 100\% & 0.301 & \textbf{0.318} & +5.6\% & 0.047 & \textbf{0.056} & +20.8\% & 0.290 & \textbf{0.306} & +5.7\% \\
 &  & 50\% & 0.226 & \textbf{0.324} & +43.2\% & 0.021 & \textbf{0.054} & +160.0\% & 0.222 & \textbf{0.310} & +39.5\% \\
 &  & 25\% & 0.047 & \textbf{0.217} & +358.2\% & 0.013 & \textbf{0.039} & +214.8\% & 0.059 & \textbf{0.225} & +284.3\% \\
 \cmidrule{2-12}
 & \multirow{3}{*}{Mistral-7B} & 100\% & 0.310 & \textbf{0.316} & +1.7\% & 0.047 & \textbf{0.048} & +1.0\% & 0.296 & \textbf{0.301} & +1.6\% \\
 &  & 50\% & 0.236 & \textbf{0.300} & +26.9\% & 0.013 & \textbf{0.043} & +244.4\% & 0.229 & \textbf{0.286} & +25.0\% \\
 &  & 25\% & 0.069 & \textbf{0.180} & +161.8\% & 0.002 & \textbf{0.019} & +900.0\% & 0.084 & \textbf{0.181} & +116.0\% \\
 \cmidrule{2-12}
 & \multirow{3}{*}{Qwen3-8B} & 100\% & 0.291 & \textbf{0.329} & +13.0\% & 0.043 & \textbf{0.064} & +49.5\% & 0.281 & \textbf{0.321} & +14.1\% \\
 &  & 50\% & 0.112 & \textbf{0.227} & +102.4\% & 0.002 & \textbf{0.038} & +1540.0\% & 0.113 & \textbf{0.228} & +102.1\% \\
 &  & 25\% & 0.027 & \textbf{0.058} & +110.6\% & 0.000 & \textbf{0.009} & - & 0.043 & \textbf{0.067} & +55.3\% \\
\midrule
\multirow{9}{*}{\rotatebox[origin=c]{90}{\textbf{RepLiQA}}} 
 & \multirow{3}{*}{Llama-3.1-8B} & 100\% & \textbf{0.655} & 0.649 & -0.9\% & 0.261 & \textbf{0.265} & +1.6\% & \textbf{0.632} & 0.629 & -0.6\% \\
 &  & 50\% & 0.314 & \textbf{0.597} & +90.0\% & 0.022 & \textbf{0.252} & +1033.7\% & 0.316 & \textbf{0.578} & +82.8\% \\
 &  & 25\% & 0.073 & \textbf{0.307} & +322.3\% & 0.008 & \textbf{0.090} & +1067.7\% & 0.124 & \textbf{0.312} & +152.3\% \\
 \cmidrule{2-12}
 & \multirow{3}{*}{Mistral-7B} & 100\% & \textbf{0.649} & 0.648 & -0.2\% & \textbf{0.265} & 0.248 & -6.4\% & 0.628 & \textbf{0.628} & +0.1\% \\
 &  & 50\% & 0.291 & \textbf{0.547} & +88.2\% & 0.023 & \textbf{0.207} & +790.3\% & 0.302 & \textbf{0.532} & +76.0\% \\
 &  & 25\% & 0.084 & \textbf{0.214} & +154.7\% & 0.001 & \textbf{0.030} & +3966.7\% & 0.121 & \textbf{0.224} & +85.8\% \\
 \cmidrule{2-12}
 & \multirow{3}{*}{Qwen3-8B} & 100\% & 0.441 & \textbf{0.478} & +8.4\% & 0.072 & \textbf{0.100} & +39.0\% & 0.420 & \textbf{0.458} & +8.9\% \\
 &  & 50\% & 0.154 & \textbf{0.349} & +126.8\% & 0.009 & \textbf{0.070} & +705.7\% & 0.150 & \textbf{0.339} & +126.5\% \\
 &  & 25\% & 0.029 & \textbf{0.065} & +126.4\% & 0.000 & \textbf{0.002} & - & 0.049 & \textbf{0.081} & +66.3\% \\
\midrule
\multirow{9}{*}{\rotatebox[origin=c]{90}{\textbf{SQuAD v2}}} 
 & \multirow{3}{*}{Llama-3.1-8B} & 100\% & 0.583 & \textbf{0.758} & +30.0\% & 0.467 & \textbf{0.658} & +40.8\% & 0.586 & \textbf{0.759} & +29.6\% \\
 &  & 50\% & 0.427 & \textbf{0.740} & +73.2\% & 0.287 & \textbf{0.638} & +122.2\% & 0.433 & \textbf{0.740} & +71.0\% \\
 &  & 25\% & 0.257 & \textbf{0.628} & +144.6\% & 0.147 & \textbf{0.483} & +229.2\% & 0.286 & \textbf{0.637} & +122.3\% \\
 \cmidrule{2-12}
 & \multirow{3}{*}{Mistral-7B} & 100\% & 0.573 & \textbf{0.750} & +30.8\% & 0.466 & \textbf{0.642} & +37.7\% & 0.576 & \textbf{0.750} & +30.3\% \\
 &  & 50\% & 0.388 & \textbf{0.654} & +68.4\% & 0.259 & \textbf{0.537} & +107.1\% & 0.398 & \textbf{0.656} & +64.8\% \\
 &  & 25\% & 0.078 & \textbf{0.500} & +541.7\% & 0.027 & \textbf{0.347} & +1192.6\% & 0.097 & \textbf{0.520} & +436.2\% \\
 \cmidrule{2-12}
 & \multirow{3}{*}{Qwen3-8B} & 100\% & 0.705 & \textbf{0.728} & +3.3\% & 0.611 & \textbf{0.648} & +5.9\% & 0.705 & \textbf{0.729} & +3.4\% \\
 &  & 50\% & 0.269 & \textbf{0.627} & +132.9\% & 0.117 & \textbf{0.542} & +363.0\% & 0.265 & \textbf{0.632} & +138.2\% \\
 &  & 25\% & 0.042 & \textbf{0.284} & +575.0\% & 0.002 & \textbf{0.165} & +8175.0\% & 0.052 & \textbf{0.310} & +492.3\% \\
\midrule
\multirow{9}{*}{\rotatebox[origin=c]{90}{\textbf{TriviaQA}}} 
 & \multirow{3}{*}{Llama-3.1-8B} & 100\% & 0.328 & \textbf{0.360} & +9.8\% & 0.266 & \textbf{0.301} & +13.3\% & 0.327 & \textbf{0.359} & +9.8\% \\
 &  & 50\% & 0.290 & \textbf{0.351} & +20.8\% & 0.249 & \textbf{0.298} & +19.5\% & 0.290 & \textbf{0.349} & +20.5\% \\
 &  & 25\% & 0.137 & \textbf{0.291} & +112.2\% & 0.098 & \textbf{0.243} & +149.2\% & 0.138 & \textbf{0.290} & +110.4\% \\
 \cmidrule{2-12}
 & \multirow{3}{*}{Mistral-7B} & 100\% & 0.205 & \textbf{0.216} & +5.6\% & \textbf{0.119} & 0.110 & -7.7\% & 0.206 & \textbf{0.216} & +4.5\% \\
 &  & 50\% & 0.138 & \textbf{0.158} & +14.6\% & 0.066 & \textbf{0.088} & +33.6\% & 0.139 & \textbf{0.157} & +13.6\% \\
 &  & 25\% & 0.032 & \textbf{0.121} & +273.9\% & 0.016 & \textbf{0.070} & +342.9\% & 0.034 & \textbf{0.123} & +257.3\% \\
 \cmidrule{2-12}
 & \multirow{3}{*}{Qwen3-8B} & 100\% & 0.146 & \textbf{0.295} & +101.7\% & 0.079 & \textbf{0.235} & +196.8\% & 0.145 & \textbf{0.295} & +102.6\% \\
 &  & 50\% & 0.032 & \textbf{0.248} & +671.9\% & 0.005 & \textbf{0.200} & +3709.5\% & 0.028 & \textbf{0.248} & +776.5\% \\
 &  & 25\% & 0.005 & \textbf{0.131} & +2287.3\% & 0.000 & \textbf{0.091} & - & 0.005 & \textbf{0.131} & +2279.9\% \\
\bottomrule
\end{tabular}%
}
\end{table}

\end{document}